# QUESTION ANALYSIS FOR ARABIC QUESTION ANSWERING SYSTEMS

Waheeb Ahmed[1], Dr. Babu Anto P[2]

[1]Research Scholar, Department of IT, Kannur University, Kerala, India
[2]Associate Professor, Department of IT, Kannur University, Kerala, India

## ABSTRACT

*The first step of processing a question in Question Answering(QA) Systems is to carry out a detailed analysis of the question for the purpose of determining what it is asking for and how to perfectly approach answering it. Our Question analysis uses several techniques to analyze any question given in natural language: a Stanford POS Tagger & parser for Arabic language, a named entity recognizer, tokenizer, Stop-word removal, Question expansion, Question classification and Question focus extraction components. We employ numerous detection rules and trained classifier using features from this analysis to detect important elements of the question, including: 1) the portion of the question that is a referring to the answer (the focus); 2) different terms in the question that identify what type of entity is being asked for (the lexical answer types); 3) Question expansion ; 4) a process of classifying the question into one or more of several and different types; and We describe how these elements are identified and evaluate the effect of accurate detection on our question-answering system using the Mean Reciprocal Rank(MRR) accuracy measure.*

## 1. INTRODUCTION

Question analysis is the first stage of any QA system and the accuracy of its results significantly impacts on the following stages of information retrieval and answer extraction. To get a better result, the semantic information available in questions should be extracted for question analysis. The question answering process in most of question-answering systems, starts with a question analysis phase that tries to determine what the question is looking for and how to effectively approach answering it[1]. Generally speaking, question analysis module receives the unstructured text question as input and identifies the syntactical and semantically elements of the question, which are kept as structured information that is used later by the many components of our QA system. Almost all of our QA system components rely in some way on the information generated by question analysis stage[2]. Question analysis is built on the top of parsing, tagging and semantic analysis components. we employ numerous recognition rules and classifiers to identify numerous critical elements of the question. There are a several and variety of such elements, each of which is crucial to different parts of the question processing phase. The most important elements are the focus, answer types (AT), Question Classification, and Question Terms(QTerms). In addition, question is expanded by adding synonyms of its terms to improve the accuracy of the retrieval process. After the question pre-processing and processing steps are done, the final stage is to extract the answer from the retrieved documents.

## 2. QUESTION ANALYSIS MODULE

This module is responsible for analyzing the question carefully before sending it to the Information Retrieval Module. The Question processing module consists of three sub-modules, the Tokenizer, Class Extractor and Focus detector as illustrated in Figure 1. The first module is for Splitting the question into individual tokens, the second module is for identifying the class of the question and the third module for extracting the question's focus. The focus of the question specifies what the given question is exactly looking for. The following figure shows the architecture of our proposed QA system along with the different sub-modules used for question processing.

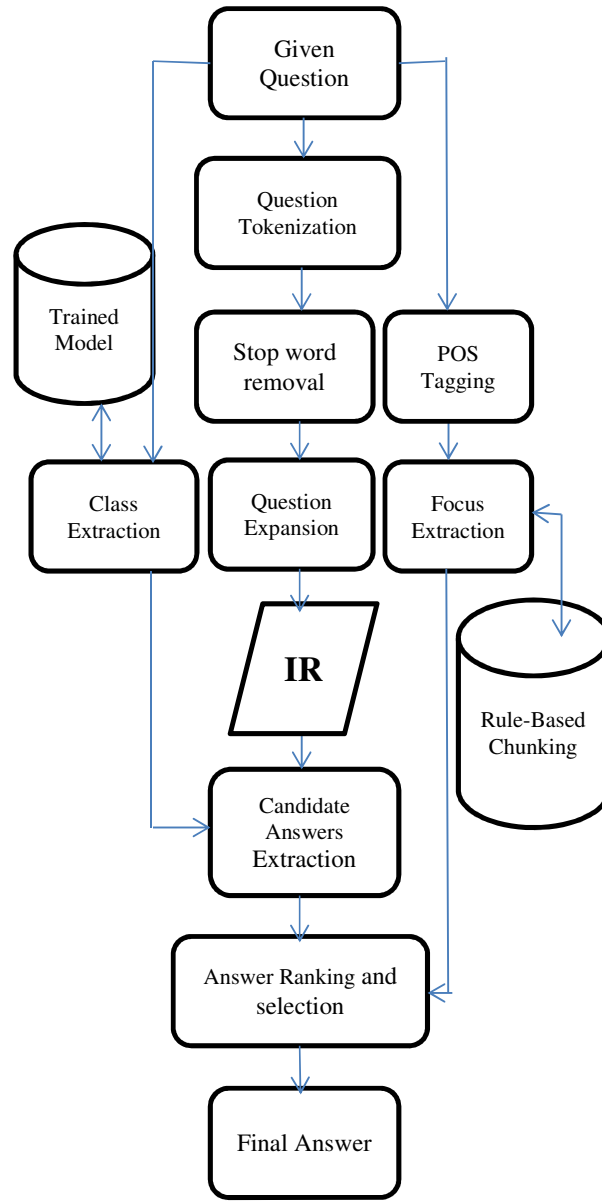

Figure 1. Question Analysis Module

Figure 1. shows the different stages through which the question is handled until the final answer is extracted and generated to the user. The following sections elaborate on the various subtasks applied to the question to extract relevant information that could assist the subsequent stages of the QA system.

## 2.1. Question tokenization

The pre-processing step is tokenization. The first step in question analysis is to identify tokens, or those elementary units which do not require to be decomposed in a subsequent processing. The entity word is defined as one kind of token for Natural Language Processing(NLP) in general and specifically in QA, the most basic one[3]. Tokenization is a crucial step in QA . It can be considered as a preparation stage for all other natural language processing tasks. Tokenization is the task of splitting words (morphemes) from running text [4]. Word Segmentation(tokenization) is getting words from text. The space is a good separator for this purpose but it will not work with special cases as compound words[5]. Some compound words are written with a space in the middle even though they are single words. Therefore, tokenization is a necessary and non-trivial step in natural language processing [6]. It is much related to the morphological analysis but usually it has been considered as an independent process [7]. Arabic words are often ambiguous in their morphological analysis. This is due to Arabic's rich language of affixes and clitics and the elimination of disambiguating short vowels and other diacritics in standard orthography ("undiacritized orthography"). On average, a word form in the Arabic Tree Bank(ATB) has about 2 morphological analyses [8]. Arabic word can come in form [Proclitics] + [inflected word] +[Enclitics]. Then, tokenization is similar/equivalent to word segmentation in Chinese language where Arabic word is as a sentence in Chinese language[9]. This sub-module splits the question into separate terms so that it can be further processed by subsequent modules in the QA system. For example, the question: "ما هي الكارثة الأكثر كلفة والتي واجهت سوق التأمين؟"("What is considered the costliest disaster the insurance industry has ever faced? ") will be split into the following tokens(؟, ما, هي, ,الكارثة, الأكثر, كلفة, و, التي, واجهت, سوق, التأمين).

## 2.2. Stop Words Removal

This sub-module removes the prepositions, Conjunctions and interrogative words. Since the prepositions and conjunctions occurs very frequently in the documents, these words can add any benefit for the information retrieval IR) module[8ooo][10]. The IR module identifies the target documents by means of the terms that are occurring very less times in the documents. After removing the stop words the remaining thing will be the important terms in the question.

## 2.3. Question Expansion

Traditional keyword based search for information is proved to have some limitations. This include word sense ambiguity, and the question intent ambiguity which can badly affect the precision. To get rid of these limitations we need to adopt semantic information retrieval techniques. These techniques are concentrating on the meaning the user looking for rather than the exact words of the user's question. We consider four main features that make users prefer semantic based search systems over keyword-based: Handling Generalizations, Handling Morphological Variants, Handling Concept matches, and Handling synonyms with the correct sense (Word Sense Disambiguation)[11][12]. In question expansion, synonyms for nouns and adjectives in the question are added to the list of question terms. Since the documents which may contain the answer for the question may not contain the terms that the user used in his question. Therefore, expanding the user question by adding synonyms to the nouns and adjectives of the question will increase the chance of getting the answer[13] and for this we used the Arabic WordNet[14].

## 2.4. Class Extraction

We used a trained Support Vector Machine(VSM) Classifier from our previous work[15]. The classifier will receive the question and give a label to the question. It is trained to produce a label based on two level classification. For example, "Why do heavier objects travel downhill faster ?" the output of the classifier will be "DESCRIPTON:reason" that is, the question is asking for descriptive answer and this is the coarse grain type of the answer. The fine grain type of the answer is "reason". The class extraction module sends its output to Answer Extraction(AE) module to apply the proper technique for extracting the answer. Table 1 shows the different classes as per the proposed scheme by Li & Roth[16].

Table 1.  Question classes

| Question class(Level 1) | Question class(Level 2) |
|---|---|
| HUMAN | Group<br>Individual<br>Title<br>Description |
| LOCATION | Country<br>State<br>City<br>Mountain<br>other |
| NUMERIC | Count<br>Date<br>Money<br>Distance<br>Speed<br>Percent<br>Other |
| DESCRIPTION | Definition<br>Manner<br>Reason |
| ENTITY | Color<br>Animal<br>Technique<br>Planet<br>other |

For example:

Question 1: "ما هي قناة جذر الأسنان؟"
("What is a dental root canal ?" )
Question Class=DESCRIPTION:definition

The general type of  answer for  question 1 is "DESCRIPTION" that is the question is looking for description and the type of description is "definition".

Question 2: "كم عدد الأشهر التي يحتاجها القمر حتى يدور حول الشمس؟"
("How many months does it take the moon to revolve around the Earth ?")
Question Class= NUMBER:count

In question 2, the answer type is "NUMBER" and more specifically a "count". Hence, the number of months(count) the moon take to revolve around the earth is the required answer.

## 2.5. Focus identification

The question focus is the set of nouns and noun phrases(NPs) available in the question. The question focus information is used by the AE module for ranking the candidate answers.

For example:

Question 3: "من هو أول أمريكي صعد الفضاء؟"
("Who was the first American in space?")
Class Extraction: HUMAN: individual
FOCUS=" أول أمريكي صعد الفضاء " (the first American in space)
FOCUS-HEAD = "أمريكي"(American)
FOCUS-MODIFIERS=ADJ "أول"(first), COMP "صعد الفضاء " (in space)

Question 4: "ما هي التقنية التي تستخدم لاكتشاف العيوب الخلقية؟"
("Name a technique widely used to detect birth defects ?")
Class Extraction: ENTITY:technique
FOCUS="التقنية التي تستخدم لاكتشاف العيوب الخلقية"(a technique widely used to detect birth defects)
FOCUS-HEAD ="التقنية"(technique)
FOCUS-MODIFIERS=ADV "بشكل واسع"(widely), COMP "التي تستخدم لاكتشاف العيوب الخلقية"(used to detect birth defects)

To extract the above information we Once the question terms are tagged , the focus, focus-head, and modifiers of the focus head could be extracted. The FOCUS chunk is extracted by rule-based technique. Where several grammar rules are applied and that is because the noun phrases can come in a variety of forms and for each form a unique grammar rule is used. The rule-based chunking for nouns and noun phrases is based on the POS Tagging information produced by Stanford POS Tagger for Arabic[17].

Question 5: "ما الدولتين الأوروبيتين اللتان دخلتا في حرب الاستقلال الأمريكية ضد البريطانيين؟ "
(What two European countries entered the War of American Independence against the British?)
Class Extraction: LOCATION:country

The POS tags foe question 5:

ما/WP الدولتين/DTNNS الأوروبيتين/DTJJ اللتان/WP دخلتا/VBD في/IN حرب/NN الاستقلال/DTNN
الامريكية/DTJJ ضد/NN البريطانيين/DTNNS

The generated syntax tree for question 5:

```
(ROOT
  (SBARQ
    (WHNP (WP ما))
    (S
      (NP
        (NP (DTJJ الدولتين) (DTNNS الاوروبيتين))
        (SBAR
          (WHNP (WP اللتان))
          (S
            (VP (VBD دخلتا)
              (PP (IN في)
                (NP
                  (NP (NN حرب)
                    (NP (DTJJ الاستقلال) (DTNN الامريكية)))
                  (NP (NN ضد)
                    (NP (DTNNS البريطانيين))))))))))))))
```

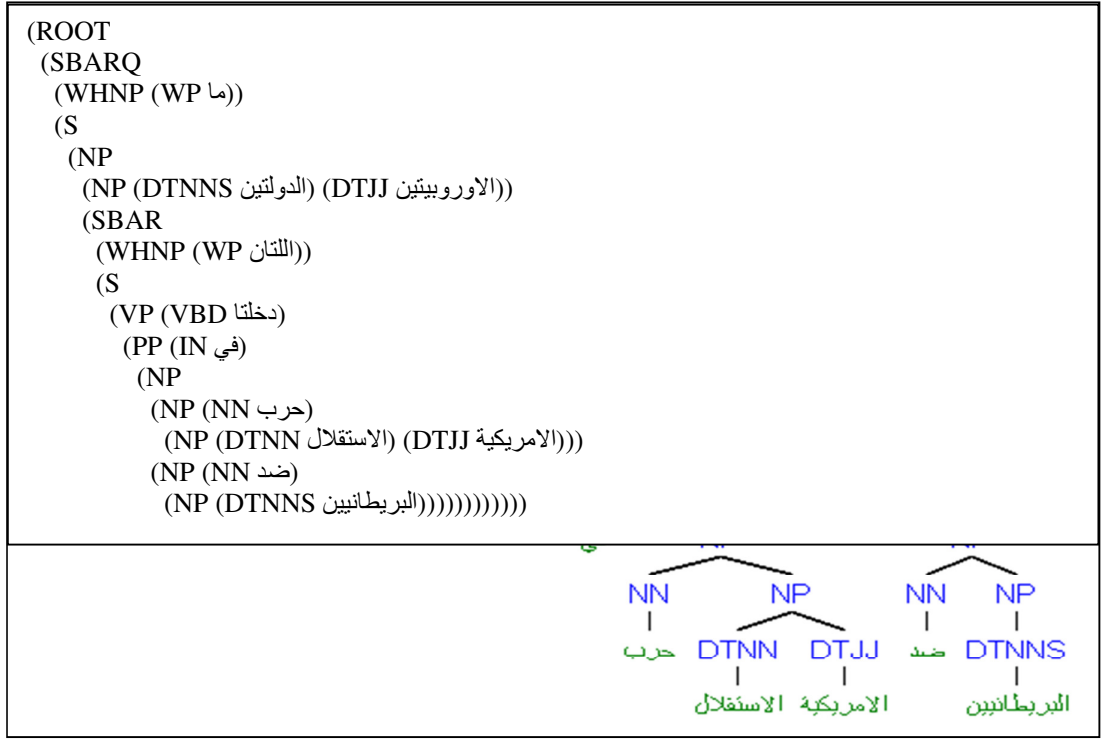

Figure 2.  Parse tree for question 5

Figure 2 shows the different noun phrases identified by NP for question 5. These noun phrases are extracted as question focus. For parsing Arabic questions, Stanford Parser for Arabic language is used[18].

## 3. DOCUMENT RETRIEVAL

The expanded list of terms extracted from the question along with the synonyms will be sent to the IR module for document retrieval, We implemented our IR module using the Vector Space Model for its simplicity of implementation and also its efficiency[19]. The system first extracts text from the  top 10 retrieved documents from which the top three documents are selected for further processing by the AE module.

## 4. ANSWER EXTRACTION

It initiates by processing a document using several procedures: first, the raw text of the document  is divided into  sentences  with the help of  a sentence segmenter, and each sentence is further subdivided into words(tokens) using a tokenizer. Next, each sentence is tagged with part-of-speech tags, which will help the named entity detection[20]. This module applies different techniques for extracting different types of answers. For example, if the question class given by the class extraction module is "HUMAN:individual" this means the question is looking for a person name. So, the AE module will use Named Entity Recognizer technique to get the answer. Questions which ask for dates a pattern matching technique will be used. Answer selection and ranking: To select answer from the top 5 generated answers/sentences by the AE module  the Answer selection and ranking stage use the question focus for this purpose.

- For extracting answer types of "HUMAN", "LOCATION" we use Named Entity Recognizer. A Named Entity Recognition (NER) system is a significant tool in natural language processing (NLP) research since it allows identification of proper nouns in open-domain (i.e., unstructured) text. For the most part, such a system is simply recognizing instances of linguistic patterns and collating them[21]. An important component of a QA system is the named entity recognizer and virtually every QA system incorporates one. Many natural language processing applications require finding named entities (NEs) in textual documents. NEs can be, for example, person or company names, dates and times, and distances. The task of identifying these in a text is called named entity recognition and is performed by a named entity recognizer (NER). The rationale of incorporating a NER as a module in a QA system is that many fact-based answers to questions are entities that can be detected by a NER. Therefore, by incorporating in the QA system a NER, the task of finding some of the answers is simplified considerably.[22] The positive impact of NE recognition in QA is widely acknowledged and there are studies that confirm it The positive impact of NE recognition in QA is widely acknowledged and there are studies that confirm it [23].

- For extracting answer types of "NUMERIC" we use Regular Expressions(RE) [24]. where a set of regular expressions for different numeric formats are used.

- For extraction answer types of "DESCRIPTION" we use semantic similarity measure between the question terms and the document sentences[25]. We developed answer extraction and passage retrieval techniques for Arabic language in our previous works[26][27]. In order to identify the relevance of a likely answer to a question, a semantic similarity calculation was employed to compute the semantic similarity between the question sentence and the answer sentence.

## 5. RESULTS AND EVALUATION

In this study, we presented a combination of techniques to question analysis, employed in a closed-domain question answering system for Arabic language. Our question analysis module consists of several subtasks importantly focus extraction and question classification. For focus extraction, we have multiple rule-based approach based on the output of Stanford POS Tagger for Arabic. Additionally, we described a classification approach for question classification. For question classification, we employed a machine learning classifier which uses a trained model to each class. In addition to the methodology presented, we also used a set of manually annotated questions for testing the system.

The assessment methods of answer extraction for the different types of questions supplied to the QA system is based on Text Retrieval Conference(TREC)[28], using MRR (Mean Reciprocal Rank) standards shown in the following formula:

$$MRR = \frac{1}{n} \sum_{i=1}^{n} \frac{1}{r_i} \qquad (1)$$

Where, n refers to the number of the questions to be tested and $r_i$ refers to the position of the  first correct answer to the question number i, if there is no correct answer available  in candidate sentences, the value will be 0. We used a set of 250 questions translated from TREC-10

Dataset. For each question type a set of 50 questions is used. The corpus is Open-domain(based the world wide web).

Table 2.  Evaluation results

| Question Type | Number | MRR |
|---|---|---|
| HUMAN | 50 | .78 |
| NUMERIC | 50 | .62 |
| LOCATION | 50 | .73 |
| ENTITY | 50 | .56 |
| DESCRIPTION | 50 | .54 |
| AVERAGE | 50 | .65 |

From table 1. It is clear that the performance of our QA system has got highest score for questions of type "HUMAN", e.g., "who is the director of NASA foundation?". The system  do well in analysing this kind of questions and this indicates the accuracy of the named entity recognizer. The system got low score for questions of type "DESCRIPTION" as this kind of questions look for descriptive answers like reason and manner and it requires more sophisticated techniques at the answer extraction module. As, there several techniques for measuring the similarity between question terms and every sentence in the top returned passages. Also, the variation of the length of the required answer, as some questions requires one sentence answer while some other questions requires 2 or 3 sentences and others full paragraph or passage. And the technique fails sometimes to return the complete answer and this decreases the accuracy value. However, the overall accuracy of the QA system is 65% which is a promising result achieved by an open domain system.

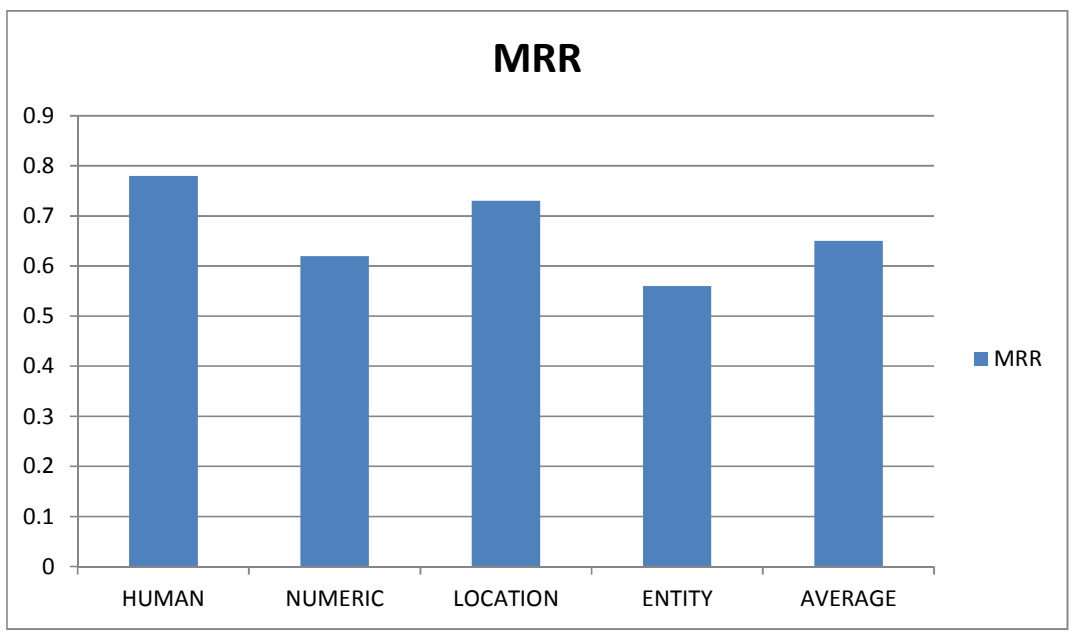

Figure 3. Distribution of the MRR for the five types of questions along with the average value

From figure 3. The average MRR achieved is 65%. Different values of MRR for each kind of question is because the analysis requires different amount of information for each kind and the complexity of some questions also requires special handling techniques.

# 6. CONCLUSION

In this paper, we have developed a question analysis module for analyzing a natural language question. Our Question analysis module is mainly concerned with the identification of four important factors , namely, focus, question expansion, Question Classification, and Q terms extraction . This is a comprehensive analysis of question which extracts all the necessary information that will be used as inputs for the other question answering components. We have evaluated our implementation of our module in terms of its performance based  on the  focus identification  and Question Classification tasks, by evaluating  its impact on our QA system accuracy. Our proposed  method achieved average accuracy of 65% for the five types of questions with total 250 questions submitted to the system.

**Authors**

**Waheeb Ahmed** received his B.E. in Computer Science & Engineering from University of Aden, Yemen, 2008. He received his master's degree from Osmania University, Hyderabad, 2013, under the ICCR-CE Programme. He is currently pursuing his Ph.D. in IT from Kannur University under the ICCR-GSS Programme.

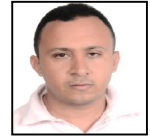

**Dr. Babu Anto P** received his Ph.D. in Technology from Cochin University of Science & Technology, India, 1992. He is currently an associate professor and working as a research guide at the Department of Information Technology, Kannur University, Kerala, India.

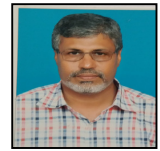